\documentclass[10pt,twocolumn,letterpaper]{article}

\usepackage[pagenumbers]{cvpr} 

\usepackage{subcaption}
\usepackage{graphicx}
\usepackage{amsmath}
\usepackage{amssymb}
\usepackage{booktabs}
\usepackage{algorithm}
\usepackage{algorithmic}

\usepackage[pagebackref,breaklinks,colorlinks]{hyperref}

\usepackage[capitalize]{cleveref}
\crefname{section}{Sec.}{Secs.}
\Crefname{section}{Section}{Sections}
\Crefname{table}{Table}{Tables}
\crefname{table}{Tab.}{Tabs.}

\def\cvprPaperID{4024} 
\def\confName{CVPR}
\def\confYear{2022}

\begin{document}

\title{Topology-Preserved Human Reconstruction with Details}

\author{
Lixiang Lin ~ Jianke Zhu\\
{\tt\small lxlin@zju.edu.cn ~ jkzhu@zju.edu.cn}\\
ZheJiang University
}
\maketitle

\begin{abstract}
It is challenging to directly estimate the human geometry from a single image due to the high diversity and complexity of body shapes with the various clothing styles. Most of model-based approaches are limited to predict the shape and pose of a minimally clothed body with over-smoothing surface. While capturing the fine detailed geometries, the model-free methods are lack of the fixed mesh topology. To address these issues, we propose a novel topology-preserved human reconstruction approach by bridging the gap between model-based and model-free human reconstruction. We present an end-to-end neural network that simultaneously predicts the pixel-aligned implicit surface and an explicit mesh model built by graph convolutional neural network. Experiments on DeepHuman and our collected dataset showed that our approach is effective. The code will be made publicly available.
\end{abstract}

\section{Introduction}
\label{sec:intro}

Human reconstruction has been studied for decades, which is essential to a large amount of real-world applications, including motion capture, digital entertainments, etc. 

Generally, it is challenging to directly estimate the geometry of human from a single RGB image due to the high diversity and complexity of body shapes. Moreover, the sophisticated clothing styles often lead to the extra difficulties. 

To tackle this critical problem, the statistical human models like SCAPE~\cite{DBLP:journals/tog/SCAPE05} and SMPL~\cite{DBLP:journals/tog/SMPL15} are proposed to reduce the searching space through the parametric models built by Principal Component Analysis~(PCA) and blend skinning. Recently, the deep neural network-based methods~\cite{DBLP:conf/cvpr/HMR18, DBLP:conf/iccv/SPIN19} try to estimate the model parameters from image without resorting to the time-consuming nonlinear optimization. Although having achieved the promising results, most of these approaches are still limited to capture the shape and pose of a minimally clothed body with over-smoothing surface, which is lack of the capability to represent a human with usual clothing and details. In spite of some parametric clothing models~\cite{DBLP:conf/3dim/NeophytouH14, DBLP:conf/eccv/DeepWrinkles18, DBLP:conf/eccv/YangFHW18}, they may not generalize well in the real-world scenario.


\begin{figure}[t]
    \centering
    \begin{subfigure}{0.10\textwidth}
        \includegraphics[width=\textwidth,trim=100 0 100 30,clip]{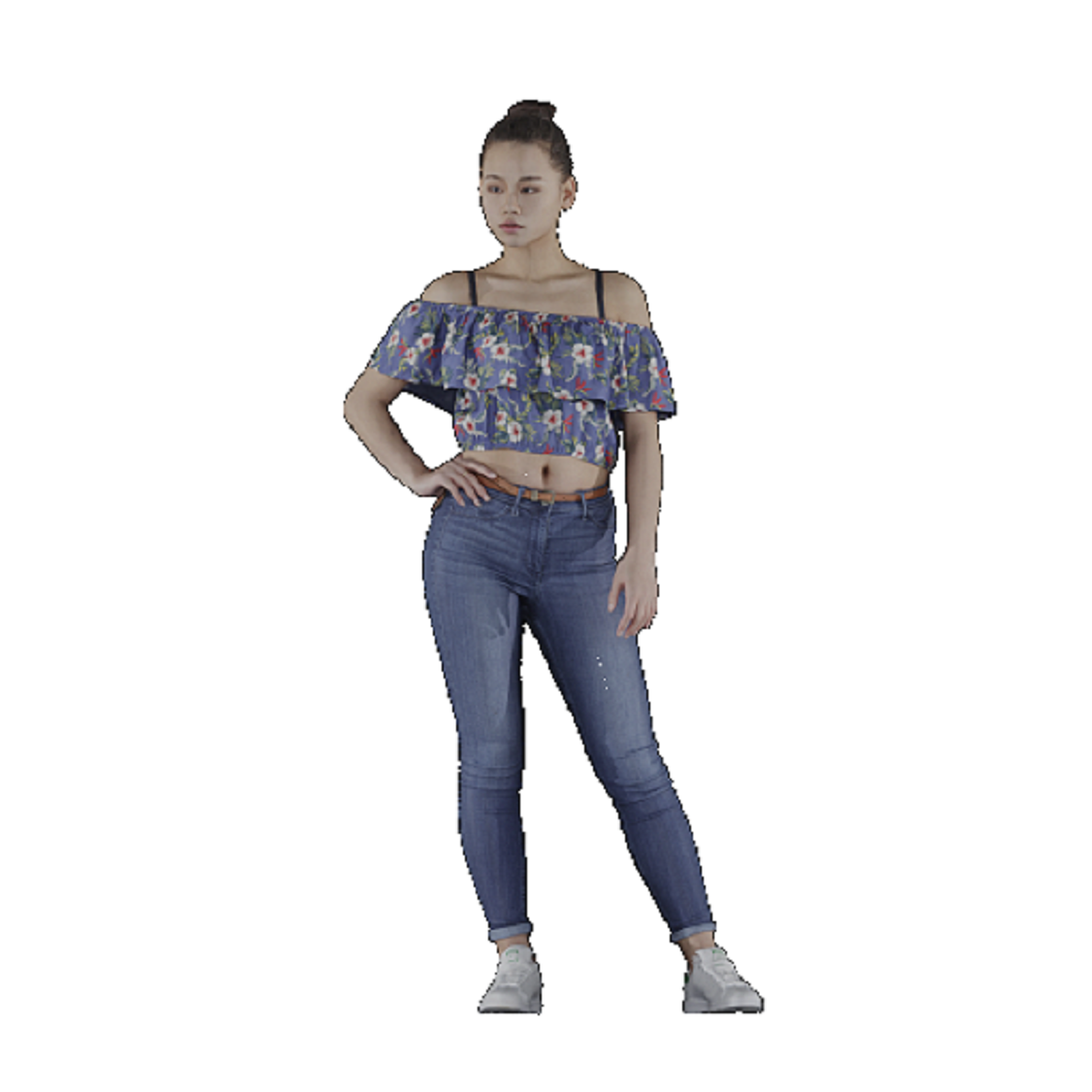}
        \caption{\scriptsize{Input image}}
    \end{subfigure}
    \hspace{-0.16cm}
    \begin{subfigure}{0.12\textwidth}
        \includegraphics[width=\textwidth,trim=100 0 100 30,clip]{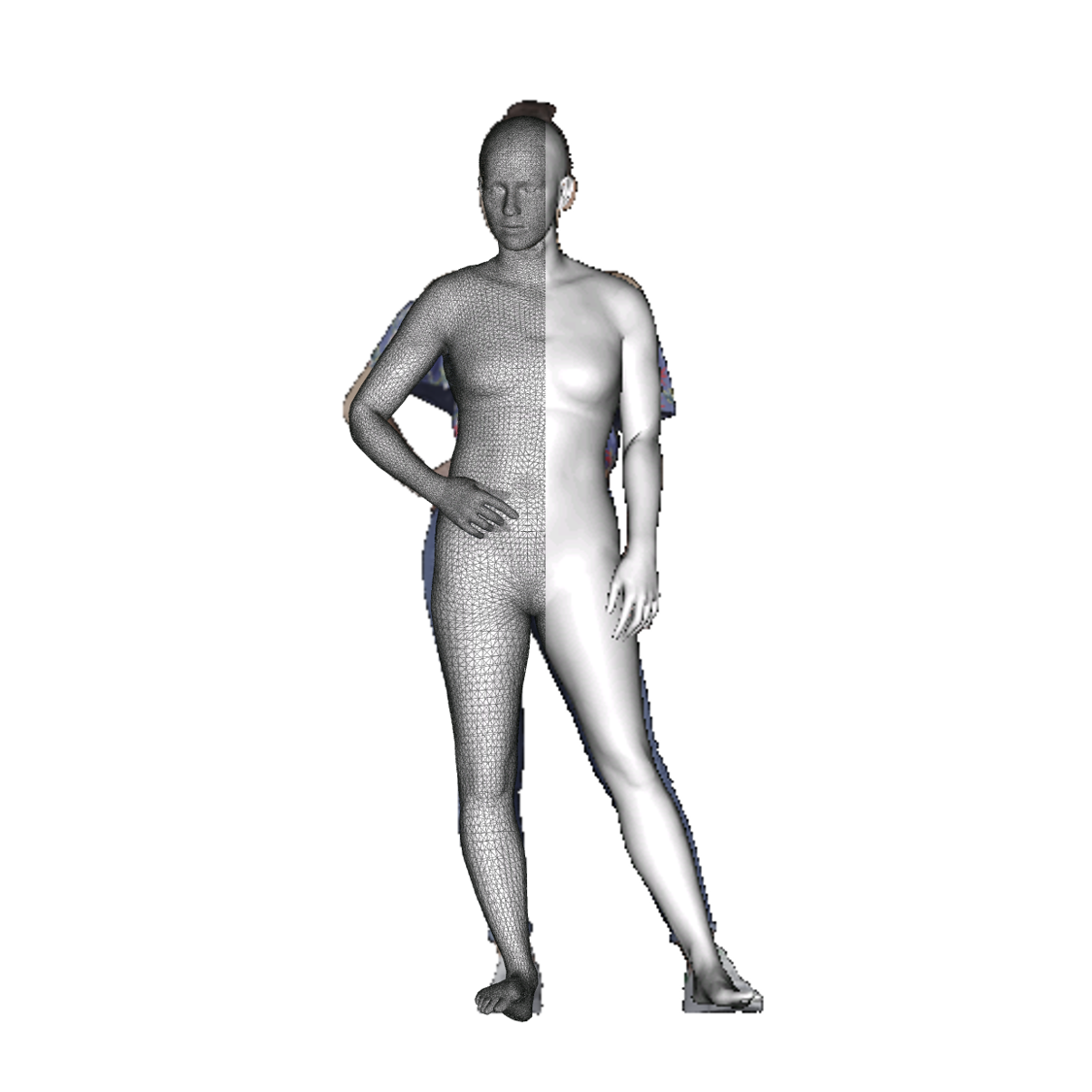}
        \caption{\scriptsize{SMPLX}}
    \end{subfigure}
    \hspace{-0.16cm}
    \begin{subfigure}{0.12\textwidth}
        \includegraphics[width=\textwidth,trim=100 0 100 30,clip]{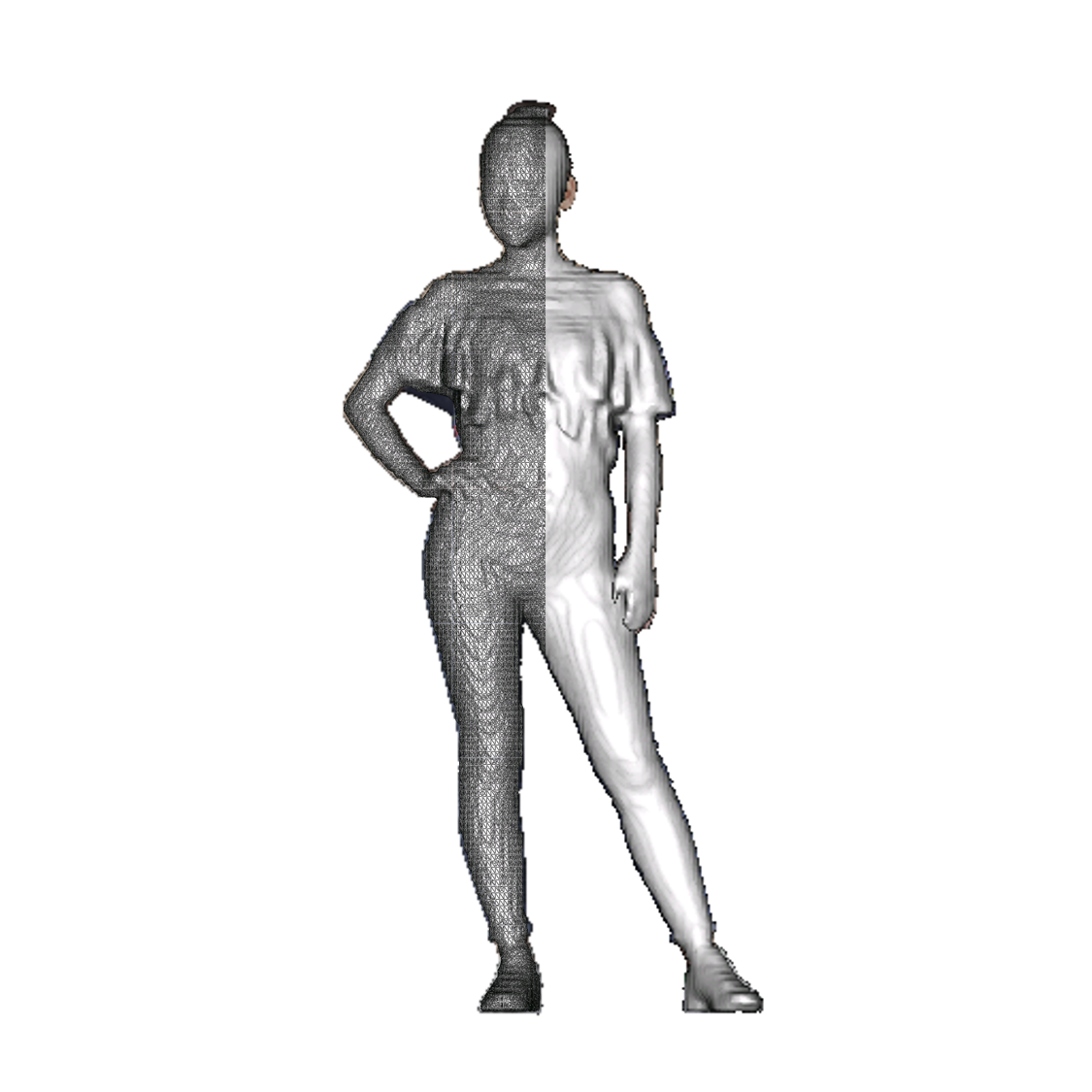}
        \caption{\scriptsize{implicit function}}
    \end{subfigure}
    \hspace{-0.16cm}
    \begin{subfigure}{0.12\textwidth}
        \includegraphics[width=\textwidth,trim=100 0 100 30,clip]{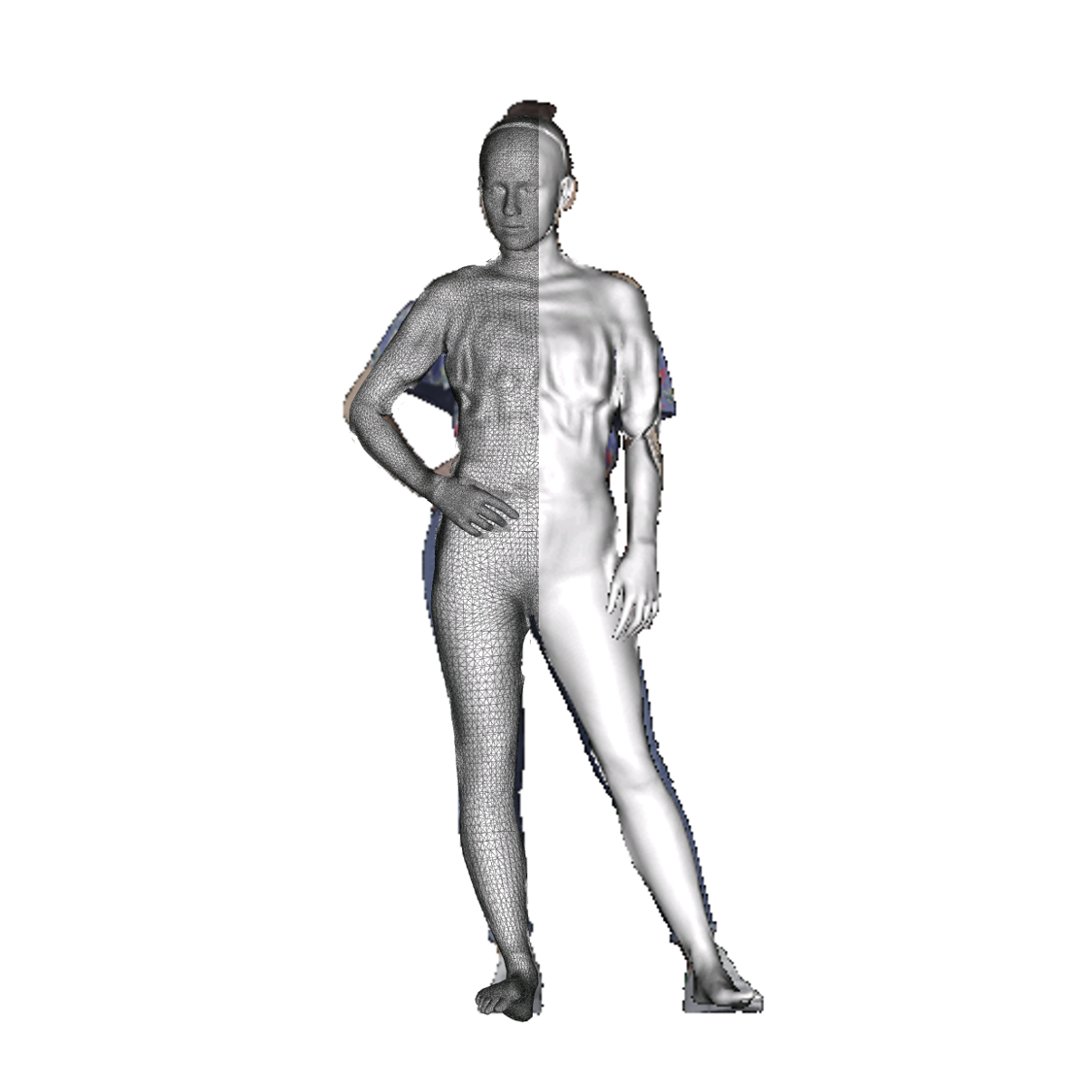}
        \caption{\scriptsize{Our method}}
    \end{subfigure}
    \hspace{-0.16cm}
    \caption{(b) Model-based approach~\cite{DBLP:conf/cvpr/SMPLX19} tries to estimate SMPLX parameters, which mainly captures the shape and pose without the details like clothing. (c) Model-free method~\cite{DBLP:conf/cvpr/pifuhd20} recovers the fine detailed body geometry while the reconstructed mesh does not have the predefined topology. (d) Our approach can directly estimate the accurate body mesh with the fixed topology.}
    \label{fig:teaser}
    \vspace{-0.2in}
\end{figure}

Instead of relying on the parametric models, the model-free approaches~\cite{DBLP:conf/eccv/BodyNet18, DBLP:conf/iccv/DeepHuman19} directly reconstruct the human body from a single image, which enjoy the merits of recovering the fine detailed geometries. To this end, human body is either estimated by the occupancy of small voxels~\cite{DBLP:conf/eccv/BodyNet18, DBLP:conf/iccv/DeepHuman19} or implicitly represented by a function learned by deep neural network~\cite{DBLP:conf/iccv/PIFu19}. The main showstopper for these methods is that there is no commonly-shared topology for the reconstructed body geometries. Therefore, it is difficult to find the semantic correspondences between the reconstructed mesh and human body part in contrast to the model-based approaches. This further prevents them from animating the reconstructed body directly.

To address the above limitations, this paper proposes a novel topology-preserved human reconstruction approach by taking advantage of both model-based and model-free methods. As shown in Fig.~\ref{fig:teaser}, we aim to accurately reconstruct the triangulated body mesh with the same topology as SMPLX model~\cite{DBLP:conf/cvpr/SMPLX19}. Specifically, we present an end-to-end neural network that simultaneously predicts the pixel-aligned implicit surface and the explicit mesh model built by graph convolutional neural network~(GCN). All the decoder branches shares the same feature encoder that greatly reduces the computational cost for inference. Finally, we suggest an effective implicit registration stage to refine the neural network output, which is performed in implicit space without resorting to the computational intensive Chamfer distance.

In summary, the main contributions of this paper are: (1) an end-to-end neural network that reconstructs the fine detailed body mesh while retaining the fixed topology from a single image; (2) a graph convolutional autoencoder to recover human mesh model with the fixed topology; (3) an efficient implicit registration method to refine the predicted mesh; (4) empirical evaluations on DeepHuman and our collected dataset showing promising human reconstruction results.

\section{Related Work}\label{sec:rel}

Recovering 3D human body shapes from 2D images or videos is the fundamental problem in computer vision, which has already been extensively studied for decades. Generally, most of existing approaches can be roughly divided into two categories. The first is dependent on the parametric models, which formulates the human body reconstruction as a regression problem. On the other hand, the model-free methods try to reconstruct detailed human geometry directly.

\subsection{Model-based Human Reconstruction}
Due to the high diversity and complexity of poses with various shapes, it is very challenging to build the human body models with the desired generalization capability. During past fifteen years, a surge of research efforts have been devoted to building the statistical human body models from 3D laser scans~\cite{DBLP:journals/tog/SCAPE05,DBLP:journals/tog/SMPL15,DBLP:conf/cvpr/TotalCapture18, DBLP:conf/cvpr/SMPLX19}. Loper \textit{et al}.~\cite{DBLP:journals/tog/SMPL15} build a skinned vertex-based model with the shape and pose parameters, in which the pose-dependent blend shapes are a linear function of the elements of the pose rotation matrices. This makes it easy to integrate the human body generation process into the deep neural network pipeline for back propagation. Recently, graph convolutional network becomes more and more important in dealing with non-rigid shape like face~\cite{DBLP:conf/eccv/COMA18}, which requires fewer parameters and can achieve higher accuracy compared with the parametric models. Choi \textit{et al}.~\cite{DBLP:conf/eccv/Pose2Mesh20} propose a graph convolutional network that recovers 3D human mesh from 2D human pose.

With the parametric human models, human reconstruction is reduced to the parameter estimation problem, where the coefficients and joints transformation are directly predicted from the still image. The conventional methods~\cite{DBLP:conf/eccv/Simplify16, DBLP:conf/cvpr/MonocularTotalCapture19, DBLP:conf/cvpr/SMPLX19} employ the nonlinear optimization solver to obtain the reasonable solution, which are usually computational intensive. Kanazawa \textit{et al}.~\cite{DBLP:conf/cvpr/HMR18} propose an end-to-end framework to recover the human body shape and pose by estimating SMPL parameters using only 2D joints annotations with an adversarial loss. Kolotouros \textit{et al}.~\cite{DBLP:conf/iccv/SPIN19} introduce a self-improving system which combines optimization and prediction methods. Most of these approaches only produce a naked human body, where the surfaces of clothing, hair, and other accessories are ignored.

To tackle the above problem, clothing is represented as an offset layer from the underlying body in~\cite{DBLP:conf/3dim/DetailedHumanAvatars18,DBLP:conf/cvpr/PeopleinClothing19,DBLP:conf/cvpr/HMD19,DBLP:journals/tog/ClothCap17}, which is able to change the pose and shape of the reconstruction using SMPL. Yang \textit{et al}.~\cite{DBLP:conf/eccv/YangFHW18} train a neural network to regress a PCA-based representation of clothing. Moreover, L{\"{a}}hner \textit{et al}.~\cite{DBLP:conf/eccv/DeepWrinkles18} learn a garment-specific pose-deformation model by regressing the low-frequency PCA components and high frequency normal maps. Adam model~\cite{DBLP:conf/cvpr/TotalCapture18} is clothed while the shape is very smooth and not pose-dependent. Recently, Ma \textit{et al}.~\cite{DBLP:conf/cvpr/CAPE20} present a generative 3D mesh model of clothed people, which is conditioned on both pose and clothing type. This enables the capability of drawing clothing samples to dress different body shapes in a variety of styles and poses. Corona \textit{et al}.~\cite{DBLP:conf/cvpr/smplicit21} propose a implicit model to represent different garment in a unified manner. In contrast to these methods, our proposed approach does not require to build an extra parametric model for the dressing, which is able to handle the cases without clothing as well.

Bhatnagar \textit{et al}.~\cite{DBLP:conf/eccv/ipnet20} recover human mesh from the incomplete point cloud by an implicit neural network to jointly predict the outer 3D surface of the dressed person, the inner body surface, and the semantic correspondences to the SMPL model. Saito \textit{et al}.~\cite{DBLP:conf/cvpr/scanimate21} propose an end-to-end trainable framework that takes raw 3D scans of a clothed human and turns them into an animatable avatar. Ma \textit{et al}.~\cite{DBLP:conf/cvpr/scale21} predict the articulated surface elements to dress the bodies with the realistic clothing that moves and deforms naturally even in the presence of topological changes. Although the above methods get the detailed human mesh with the fixed topology, they requires point cloud as the input comparing to the RGB images.

\begin{figure*}[t]
    \centering
    \includegraphics[width=17cm]{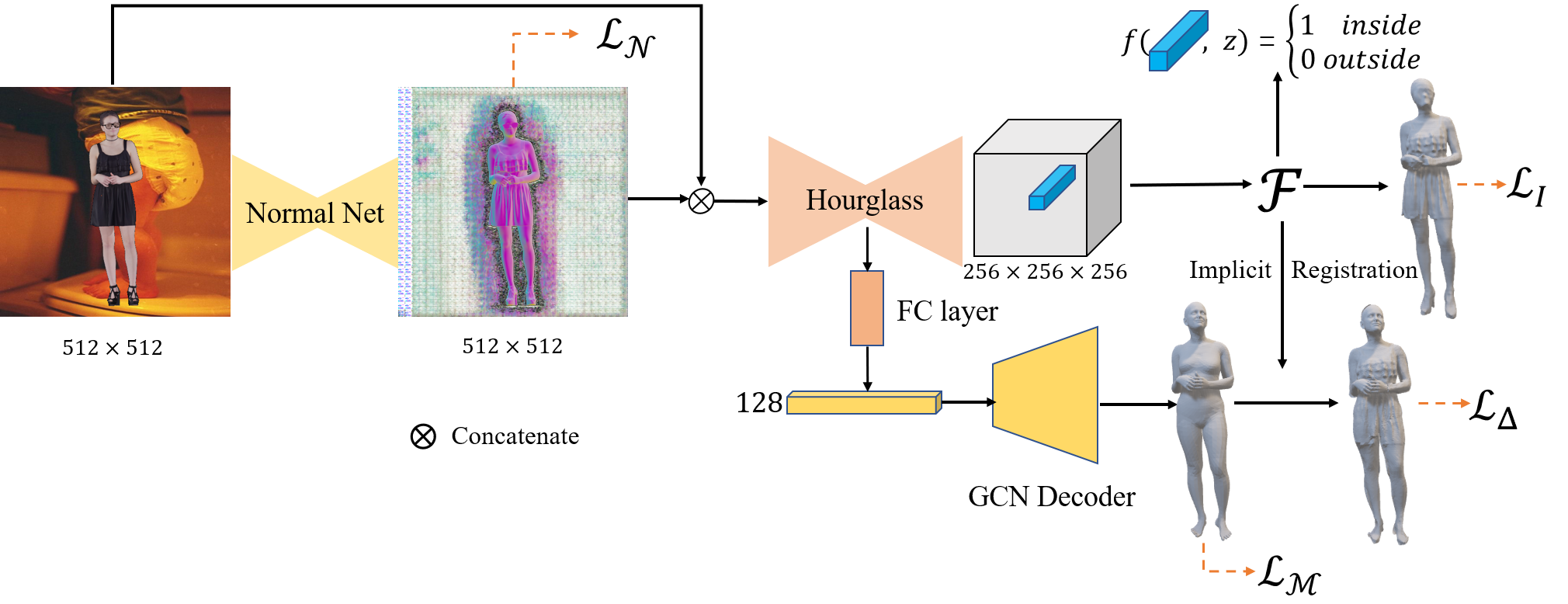}
    \caption{Overview of our proposed human reconstruction approach. We firstly concatenate the input image with the estimated normal map to feed the hourglass encoder. Then, a pixel-aligned implicit function predicts the occupancy, and a GCN decoder estimate the mesh model. Finally, the mesh model is refined through an effective implicit registration stage.}
    \label{fig:framework}
    \vspace{-0.2in}
\end{figure*}

\subsection{Model-free Human Reconstruction}
Model-free approaches try to directly estimate human body geometry like voxels or implicit surface from the still image without resorting to a prior model, which have much larger solution space to represent the fine details.

Varol \textit{et al}.~\cite{DBLP:conf/eccv/BodyNet18} suggest to learn a voxel representation of human body through the deep neural network, where the fine-scale details are often missing due to the high memory requirements of voxel representations. Zheng \textit{et al}.~\cite{DBLP:conf/iccv/DeepHuman19} introduce a discretized volumetric representation to reconstruct the detailed human, which fuses the different scales of image features in order to recover the accurate surface geometry. In spite of impressive results, the cubic memory requirement imposed by the discrete voxel representation prevents these methods from obtaining the high resolution reconstruction results. Instead of using the voxels, some approaches~\cite{DBLP:conf/iccv/MouldingHumans19,DBLP:conf/iccv/HumanDepth19} try to predict the depth maps of human as output. Natsume \textit{et al}.~\cite{DBLP:conf/cvpr/SiCloPe19} present a multi-view inference method by synthesizing silhouette views from a single image. Although multi-view silhouettes are more memory efficient, the concave regions are difficult to infer as well as the consistently generated views.
 
Saito \textit{et al}.~\cite{DBLP:conf/iccv/PIFu19} propose a memory efficient approach that represents the detailed human by a pixel-aligned implicit function. Instead of explicitly discretizing the output space into voxels, it learns a function that determines the occupancy for any locations. With such implicit representation, the occupancy for the sampled 3D point can be computed on the fly, which greatly saves the memory. Later, Saito \textit{et al}.~\cite{DBLP:conf/cvpr/pifuhd20} introduce a multi-level architecture for high-resolution 3D human reconstruction, where the coarse level focuses on the holistic reasoning and the fine level estimates the highly detailed geometry. 

\section{Methods}\label{sec:method}

In this section, we present our proposed approach to human reconstruction from a single image. Firstly, we propose the end-to-end neural framework that reconstructs the fine detailed body mesh while retaining the fixed topology. Secondly, we describe the model-free reconstruction using implicit surface loss. Thirdly, we introduce the graph convolutional network approach to recover human mesh model. Finally, we propose an effective implicit registration stage to fill the gap between the pixel-aligned implicit surface and the recovered mesh.

\subsection{Overview}
The model-based human reconstruction method~\cite{DBLP:conf/cvpr/HMR18} enjoys the merit of the predefined mesh topology, which is able to preserve the body shape through the statistical models. On the other hand, the model-free method can recover the fine detailed geometry like wrinkles on the clothing. The key idea of our proposed approach is to take advantage of both representations. To this end, we aim to reconstruct the triangulated body mesh accurately while preserving the same topology as SMPLX model. As illustrated in Fig.~\ref{fig:framework}, we present an end-to-end deep neural network with a typical encoder-decoder structure.

Our overall framework consists of four parts. Firstly, we train a Pix2PixHD network~\cite{DBLP:conf/cvpr/pix2pixHD18} with nine residual blocks and four downsampling layers to obtain the frontal normal map. Secondly, we concatenate the predicted normal map with the original image as the input for the stacked hourglass network~\cite{DBLP:conf/eccv/HourGlass16} with four stacks to extract deep features. Hourglass network produces a feature map, where we perform average pooling to get a feature vector as the input of our GCN decoder. Thirdly, a fully connected layer with the number of neurons of (1024, 1024, 1024, 128) is used to adapt the number of features. Finally, a fully connected layer with the number of neurons (257, 1024, 512, 256, 128, 1) and the skip connections at (3, 4, 5) layers are employed to predict the binary occupancy value for any given positions, and a pre-trained GCN decoder is used to get the human mesh with fixed topology.

In~\cite{DBLP:conf/cvpr/pifuhd20}, the frontal normal map is predicted as a proxy for 3D geometry. Features extracted from the frontal normal map can generate the sharper reconstructed results. Therefore, we firstly employ a Pix2PixHD network to obtain the frontal normal map, and then concatenate it with the original image as the input for the feature encoder. The Pix2PixHD network is trained with the following loss function:
\begin{equation}
    \mathcal{L_N} = \lambda_{N} \sum_{\{i,j\} \subset P}|n_{i,j} - n_{i,j}^*|
\end{equation}
where $\mathcal{L_N}$ is the regular $L_1$ loss. We try to predict the frontal normal map of the person in the image. The weight is $\lambda_{N}=5$. We use Adam optimizer with the learning rate of $2 \times 10^{-4}$ until the convergence.

It is worthy of mentioning that we suggest to share the same feature encoder for all the decoder branches. This greatly reduces the computational cost during the inference. Moreover, a decoder branch is employed to predict the implicit surface function for the model-free human reconstruction, and another decoder branch is used to extract the explicit mesh surface using graph convolutional neural network trained on a large corpus. More importantly, we propose an extra implicit registration stage to fill the gap between the other two branches, which intends to reduce the registration error between the triangulated mesh and implicit surface.

From the above all, the proposed deep neural network minimizes the following loss function:
\begin{equation}
\mathcal{L} = \mathcal{L_{I}} + \mathcal{L_{M}} + \mathcal{L}_{\Delta} 
\end{equation}
where $\mathcal{L_{I}} $ is the loss for implicit surface function estimation, and $\mathcal{L_{M}} $ is the loss to recover the human mesh through GCN decoder. $\mathcal{L}_{\Delta}$ is the loss for the implicit registration stage, which bridges the gap between the model-free reconstruction and parametric mesh model.

\subsection{Implicit Reconstruction Loss}
Motivated by the previous model-free human reconstruction approach~\cite{DBLP:conf/iccv/PIFu19}, we try to estimate the body surface through an implicit function $f(\cdot)$ that approximates the signed distance of zero level set. The implicit surface shares the same coordinate space as SMPLX mesh model. Specifically, a fully connected layer is employed to predict the binary occupancy value for any given positions $ X_i=(x_i, y_i, z_i)\in \mathbb{R}^3 $ in the continuous 3D space: 
\begin{equation}
    f(\mathcal{F}_{\mathbf{x}_i}, Z_i) = \left\{
    \begin{array}{lr}
    1, \quad \mathrm{if\  }X_i \mathrm{\ is\ inside\ mesh\ surface}\\
    0, \quad \mathrm{otherwise}
    \end{array}
    \right. \label{equ:imfun}
\end{equation}
where $\mathcal{F}_{\mathbf{x}_i}$ denotes the deep features sampled at the location $\mathbf{x}_i=(u_i,v_i)=\pi(X_i)$ in image space $\Omega \subset R^2$. The projection function $\pi:R^3\rightarrow \Omega$ can be either orthogonal projection or perspective projection. $Z_i=(MX_i)^{z}$ is the depth value in camera coordinate space, $M$ is the camera extrinsic matrix.

Given the the ground truth occupancy $y(X_i)$ at point $X_i$, we employ the extended Binary Cross Entropy (BCE) loss~\cite{DBLP:conf/iccv/DeepHuman19} to supervise our proposed implicit surface representation layer. Therefore, the implicit reconstruction loss $\mathcal{L}_{I}$ can be derived as follows:
\begin{equation}
\begin{split}
    \mathcal{L}_{I} &=  \sum_{X_i \in \mathcal{S}}\eta y(X_i)\log f(X_i)  \\ & +(1-\eta)(1-y(X_i))\log(1-f(X_i))
\end{split}
\end{equation}
where $\mathcal{S}$ is the set of the sampled points. $\eta$ represents the ratio of points outside surface in $\mathcal{S}$, which is computed from the sampling results. A mixture sampling strategy is used to select the points for the implicit reconstruction loss computation. In our experiment, the sampled points are composed of the uniform sampling and importance sampling with the standard deviations of 0.04 and the ratio of $8:1$. For the ground truth points and their occupancy, we make use of the DeepHuman dataset~\cite{DBLP:conf/iccv/DeepHuman19} and our collected high resolution human scans.

\subsection{Mesh Recovery Loss}\label{ssec:2}

The model-based human reconstruction has the merit of the watertight mesh representation with the data-driven priors, where the generative SMPLX model~\cite{DBLP:conf/cvpr/SMPLX19} is recently proposed. It has a mesh with 10,475 vertices and 54 body joints. Moreover, an extra joint is used to control the global rotation, which is parameterized by the PCA shape coefficients and poses. Although formulating the pose blend shapes as a linear function of the rotation matrices, the whole procedure of mesh generation is still highly nonlinear. Therefore, it is challenging to regress them from a single image directly.

To deal with this problem, we suggest to make use of a graph convolutional network-based autoencoder to capture human body shapes, which shares the same mesh topology as SMPLX model without the blend skinning. In this paper, we employ the same loss function described in~\cite{DBLP:conf/eccv/Pose2Mesh20} to train our proposed GCN autoencoder and regress the latent coefficients from the extracted feature. For each vertex $V_i$ in $\mathcal{M}$ with the target $V_i^*$, the mesh recovery loss $\mathcal{L_{M}}$ is defined as below:
\begin{equation}
\mathcal{L}_{\mathcal{M}} = \lambda_{v} \sum_{V_i \in \mathcal{M}}||V_i - V_i^*||_1 + \lambda_{e} \mathcal{L}_{edge} + \lambda_{n} \mathcal{L}_{normal}
\end{equation}
where the first term denotes the per vertices $L_1$ fitting loss, and the last two terms regularize the mesh deformations on edges and normals, respectively. Let $\mathcal{T}$ represent a facet in $\mathcal{M}$, and ($i,j$) are the vertex indices in $\mathcal{T}$. The edge length loss is derived as follows:   
\begin{equation}
    \mathcal{L}_{edge} = \sum_{\mathcal{T}\in \mathcal{M}} \sum_{\{i,j\} \in \mathcal{T}}|||V_i-V_j||_2 - ||V_i^*-V_j^*||_2|
\end{equation}
Given the target normal $n^*_f$ for each facet $\mathcal{T}$, the normal consistency loss is defined as in~\cite{DBLP:conf/eccv/Pose2Mesh20}:
\begin{equation}
    \mathcal{L}_{normal} = \sum_{\mathcal{T}\in \mathcal{M}} \sum_{\{i,j\} \in \mathcal{T}} \left | \left \langle \frac{V_i-V_j}{||V_i-V_j||_2}, 
    n_f^* \right \rangle  \right |
\end{equation}
The weights are $\lambda_{v}=10$, $\lambda_{e}=40$, and $\lambda_{n}=0.5$, respectively.

As in~\cite{DBLP:conf/cvpr/CAPE20}, our proposed autoencoder consists of an encoder-decoder pair built by graph convolutional network. To embed the input data into the latent space, the encoder is made of eight Chebyshev Residual Blocks with Chebyshev polynomial of order two, a Chebyshev convolution with order one and a fully connected layer.  Each graph convolution layer is followed by a Leaky ReLU~\cite{leakyrelu13}. The architecture of decoder is similar to the encoder. For the detailed network structure, please refer to the supplementary materials. To effectively capture the various body shapes and poses, we train this autoencoder on AMASS datasets~\cite{AMASS:ICCV:2019}. In contrast to COMA~\cite{DBLP:conf/eccv/COMA18} reconstructing the smoothing facial meshes, our proposed method has to tackle the critical challenges of body articulations and blend skinning.

Once GCN autoencoder is trained, we freeze the model parameters of decoder and integrate it into our proposed human reconstruction framework to facilitate the mesh model generation. Moreover, we formulate the model-based reconstruction as the GCN latent embedding estimation problem. We employ $\mathcal{L_M}$ to supervise the training process.

\subsection{Implicit Registration Loss}\label{ssec:3}

In order to take advantage of both implicit function and topology reserved human model, we propose a novel implicit registration loss to capture the detailed clothing information from implicit function. $\mathcal{L}_{sdf}$ is defined as follows:
\begin{equation}
     \mathcal{L}_{\Delta} =  \lambda_{sdf} \mathcal{L}_{sdf} + \mathcal{L}_{reg}
\end{equation}
where $\lambda_{sdf}=10$. $\mathcal{L}_{sdf}$ is defined as follows: 
\begin{equation}
    \mathcal{L}_{sdf} = \frac{1}{|\mathcal{M}|}\sum_{V_i \in \mathcal{M}} ||f(\mathcal{F}_{\pi(V_i+M^{-1}\Delta_i)}, (MV_i)^{z}+\Delta_i) - \sigma ||_1
\end{equation}
where $f(\cdot)$ is the pixel-aligned implicit function defined in Eq.~(\ref{equ:imfun}), $\Delta=(0,0,\Delta_z)$ is an optimizable variable initialized to 0, Since the learned implicit function is fed with the depth along the ray defined by the 2D projection, we only optimize it along z-axis.  $M$ is the camera extrinsic matrix, and $\sigma$ is set to $0.5$.

The regularization term $\mathcal{L}_{reg}$ is proposed to enforce the surface smoothing through minimizing the mesh Laplacian differences and the $L_2$ norm of offset $\Delta$: 
\begin{equation} 
    \mathcal{L}_{reg} = \lambda_{lap}||L(V+M^{-1}\Delta) - L(V)||_2^2 + \lambda_{norm}|| \Delta ||_2^2
\end{equation}
$L$ denotes the Laplacian matrix that retains the mesh regularity. The $L_2$ norm over mesh offset $\Delta$ prevents the vertices from shifting too large. The regularization coefficient $\lambda_{lap}$ is set to $10^4$, and $\lambda_{norm}$ is $50$.

Since the neural network prediction is close enough to the implicit surface of model-free reconstruction, $\mathcal{L}_{sdf}$ is able to guarantee the convergence. Our proposed implicit registration method does not calculate the point-to-surface Chamfer distance which is very computational intensive. Thus, the registration is performed very efficiently in implicit space without extracting the explicit mesh by the marching cube algorithm~\cite{DBLP:conf/siggraph/Marchingcubes87}.

\section{Experiments}\label{sec:exp}

In this section, we give the details of our experimental implementation and discuss the results on human reconstruction. We examine the representation capability of our proposed GCN autoencoder for human body. Moreover, we evaluate our results on DeepHuman and our collected human dataset. 

\subsection{Experimental Setup}

AMASS~\cite{AMASS:ICCV:2019} is used to train our GCN autoencoder. AMASS is a large database of human motion datasets with a common framework and parameterization. AMASS contains a large variety of SMPL and SMPLX parameters for human motion. Due to the flexibility of the face and hands, AMASS dataset does not provide the ground-truth SMPLX parameters for face and hands. With the SMPLX topology, the face and hands can be fitted with other algorithm~\cite{DBLP:conf/cvpr/SMPLX19}. Since DeepHuman dataset only provides the ground-truth SMPL parameters, we train a SMPL GCN autoencoder. To effectively optimize the GCN parameters, we use Adam optimizer with the learning rate of $10^{-4}$ and a weight decay of $10^{-4}$ for 10 epochs.

To facilitate the effective experimental evaluation, we conduct the experiments on DeepHuman dataset~\cite{DBLP:conf/iccv/DeepHuman19} and our collected scans. DeepHuman dataset contains the total number of 6,795 items, including RGB image, SMPL parameters and meshes reconstructed by a multi-view fusion algorithm. We randomly split the samples to form training and testing sets with a ratio of $9:1$, and obtain 6,115 items for training and 680 samples for testing. we crop the images according to their height and place the human at the center. Then, the cropped images are resized into the resolution of $512 \times 512$. Due to the privacy issue, the facial regions in image are blurred. Being difficult to recover the thin structures like fingers, the hand geometry of the mesh in the dataset are presented in the form of fists.

We collected 260 high-resolution photogrammetry scans on the internet. The dataset is splitted into a training set of 234 subjects and a testing set of 26 subjects and render the meshes using blender. Each subject is rendered from every $18$ degree in yaw axis with an elevation fixed with $0^{\circ}$. As in~\cite{DBLP:conf/cvpr/pifuhd20}, we randomly augment the background images using COCO dataset~\cite{DBLP:conf/eccv/coco14}. In our experiment, we render the images in the resolution of $1024 \times 1024$, and then scale it into $512 \times 512$ as the input of our network. After rendering, we employ the conventional optimization-based method~\cite{DBLP:conf/cvpr/SMPLX19} to fit the SMPLX model with respect to each scan.

We implemented the proposed approach by PyTorch. The normal estimation network is trained using Adam optimizer with the learning rate of $2 \times 10^{-4}$ until convergence. We train the pixel-aligned implicit function and deep feature for the GCN latent space encoder with 60 epochs. We use RMSprop optimizer with the learning rate $5 \times 10^{-4}$ that is decayed by the factor of 0.1 at 40-th epoch. We employ same sampling strategy as PIFu~\cite{DBLP:conf/iccv/PIFu19} to sample $12000$ points to train the implicit function. In the implicit registration stage, Adam optimizer with learning rate $0.002$ is employed to optimize the offset of mesh vertices with 500 iterations. 

\subsection{Evaluation on Autoencoder}

We evaluate the performance of our GCN autoencoder on AMASS dataset and Human3.6M dataset~\cite{DBLP:journals/pami/Human3.6M14}. Human3.6M consists of 3.6 million 3D Human poses acquired by recording the performance of 5 female and 6 male subjects.

The mean per vertex position error (MPVPE) is similar to MPJPE~\cite{DBLP:conf/cvpr/HMR18} while we make use of all vertices to evaluate the representation capability of our GCN autoencoder. As shown in Table~\ref{tab:autoencoder results}, our proposed GCN autoencoder achieves almost the same reconstruction results as the conventional parametric SMPLX and SMPL model with the fewer latent parameters.

\begin{table}[htbp]
    \centering
    \begin{tabular}{c|c}
    \toprule
        Dataset & MPVPE\,(mm) \\
    \midrule
        AMASS\,(SMPLX) & 4.628 \\
        Human3.6M\,(SMPL) & 4.913 \\
    \bottomrule
    \end{tabular}
    \caption{Evaluation on autoencoder.}
    \label{tab:autoencoder results}
\end{table}

\subsection{Evaluation on DeepHuman}

As in PIFu~\cite{DBLP:conf/iccv/PIFu19}, we adopt three reconstruction performance metrics including the mean point-to-surface Euclidean distance (P2S), Chamfer distance and normal projection error. P2S and Chamfer distance measure the reconstruction accuracy comparing to the ground-truth mesh. Additionally, the normal projection error is used to evaluate the fineness of reconstructed local details as well as the projection consistency.

\begin{figure}
    \centering
    \includegraphics[width=0.12\textwidth,trim=150 10 150 30,clip]{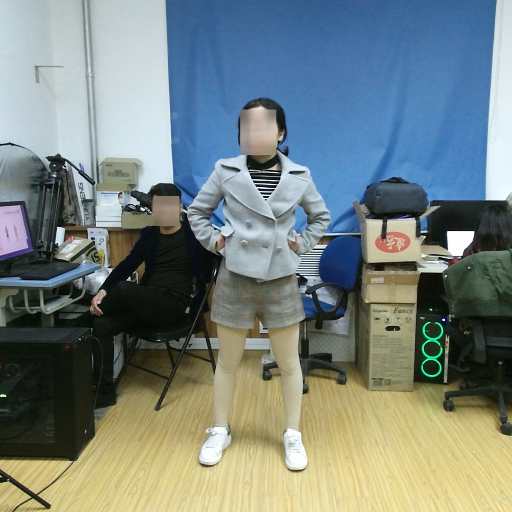}
    \hspace{-0.16cm}
    \includegraphics[width=0.12\textwidth,trim=150 10 150 30,clip]{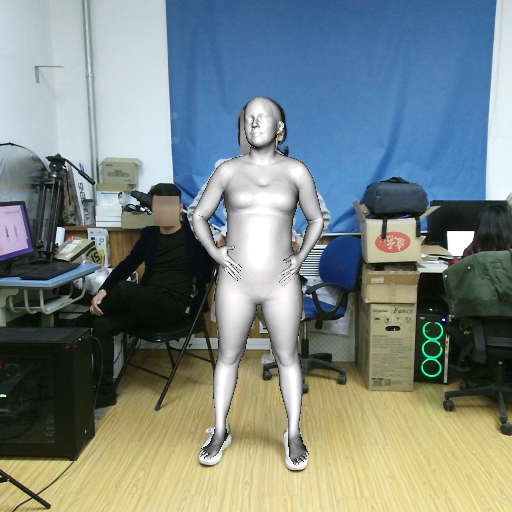}
    \hspace{-0.16cm}
    \includegraphics[width=0.12\textwidth,trim=150 10 150 30,clip]{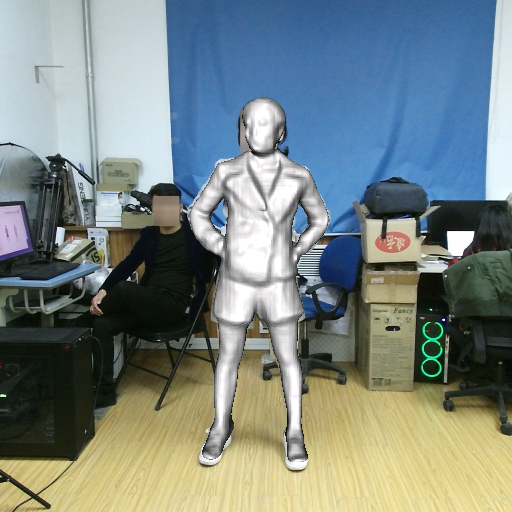}
    \hspace{-0.16cm}
    \includegraphics[width=0.12\textwidth,trim=150 10 150 30,clip]{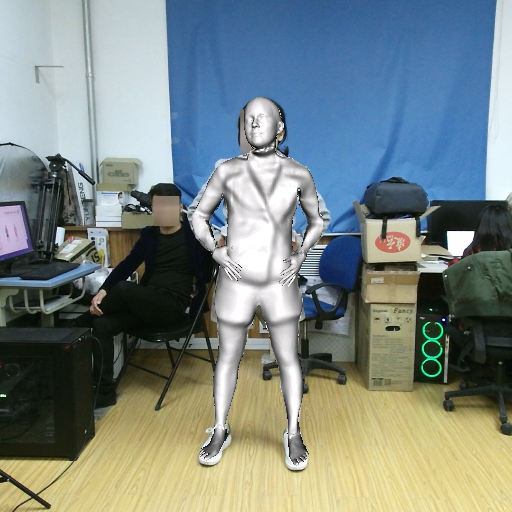}
    \hspace{-0.16cm}
    
    \begin{subfigure}{0.12\textwidth}
        \includegraphics[width=\textwidth,trim=150 10 150 30,clip]{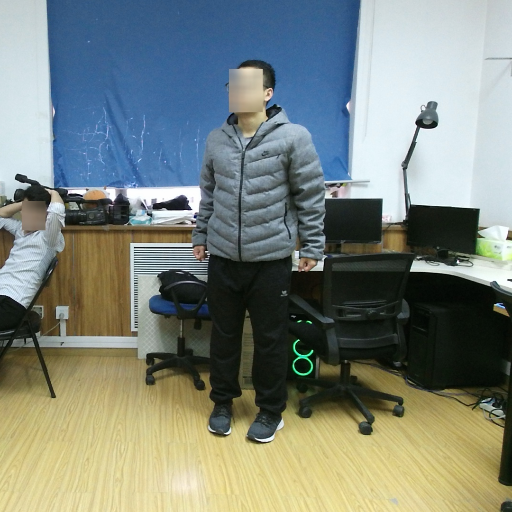}
        \caption{Input image}
    \end{subfigure}
    \hspace{-0.16cm}
    \begin{subfigure}{0.12\textwidth}
        \includegraphics[width=\textwidth,trim=150 10 150 30,clip]{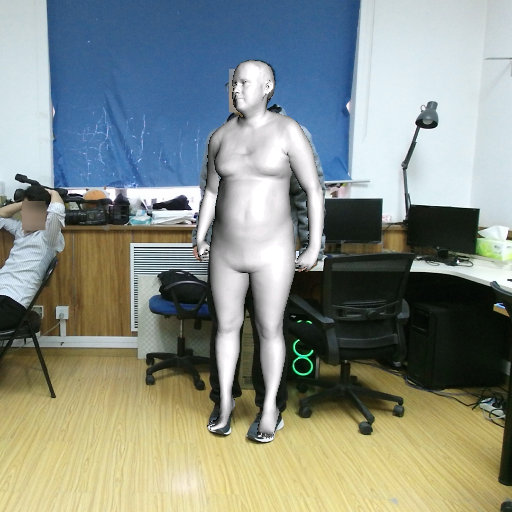}
        \caption{SMPL}
    \end{subfigure}
    \hspace{-0.16cm}
    \begin{subfigure}{0.12\textwidth}
        \includegraphics[width=\textwidth,trim=150 10 150 30,clip]{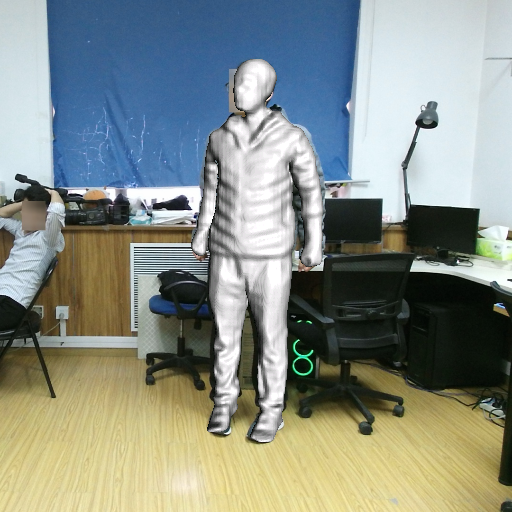}
        \caption{\scriptsize{implicit function}}
    \end{subfigure}
    \hspace{-0.16cm}
    \begin{subfigure}{0.12\textwidth}
        \includegraphics[width=\textwidth,trim=150 10 150 30,clip]{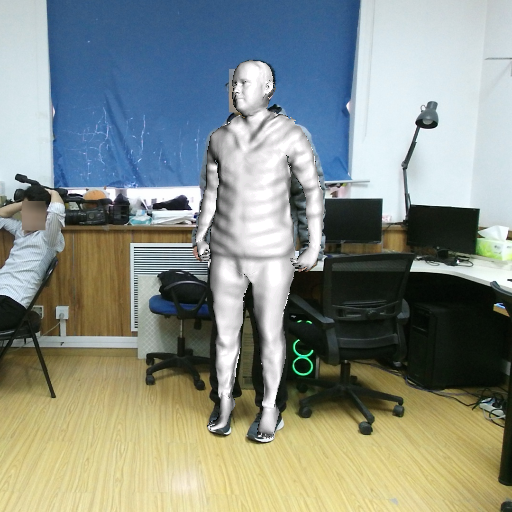}
        \caption{implicit loss}
    \end{subfigure}
    \hspace{-0.16cm}
    \caption{Reconstruction results on DeepHuman dataset. We show the results of SMPL (b), our implicit function results (c) and implicit registration results (d), respectively.}
    \label{fig:deephuamn results}
    \vspace{-0.15in}
\end{figure}
Table~\ref{tab:evaluation on deephuman} gives the experimental results on DeepHuman dataset. It can be seen that our proposed approach performs better than PIFu. Moreover, the normal map can significantly improve the reconstruction accuracy and capture the clothing details, which makes it easier for the implicit function to retain the local details.

In the implicit registration stage, we add the clothing details obtained from our trained implicit function to the SMPL mesh. The results shows that our proposed implicit registration method performs comparable with conventional Chamfer distance-based method. Qualitative results of our method are shown in Fig.~\ref{fig:deephuamn results}, GCN decoder predicts the coarse mesh with the same topology as SMPL. After implicit registration stage, the vertices offsets representing the clothing details are obtained from the implicit function. Due to the flexibility and low reconstruction quality of hands, feet and face, we do not optimize the hands, feet and face of the SMPL model.

\begin{table}
    \centering
    \begin{tabular}{c|c|c|c}
    \toprule
        Methods & Normal & P2S & Chamfer \\ 
    \midrule
        PIFu~\cite{DBLP:conf/iccv/PIFu19} & 0.020 & 2.718 & 2.327 \\
        Ours w/o normal & 0.018 & 2.413 & 2.229 \\
        Ours & \textbf{0.010} & \textbf{1.317} & \textbf{1.152} \\
    \midrule
        *Inference results & 0.030 & 1.434 & 1.278 \\
        *Refined by Chamfer loss & 0.020 & 1.324 & 1.170 \\
        *Refined by implicit loss & 0.026 & 1.335 & 1.175 \\
    \bottomrule
    \end{tabular}
    \caption{Performance evaluation on DeepHuman dataset. * indicates this output mesh has the same topology as parametric human model. Units for point-to-surface and Chamfer distance are in cm.}
    \label{tab:evaluation on deephuman}
    \vspace{-0.2in}
\end{table}

\begin{figure}[htbp]
	\centering
	\includegraphics[width=0.12\textwidth,trim=150 10 150 30,clip]{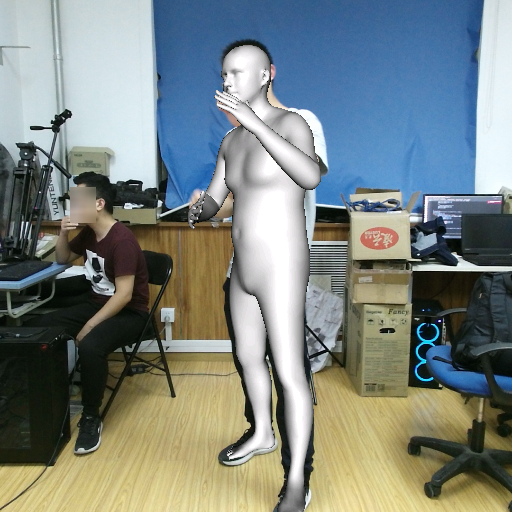}
	\hspace{-0.16cm}
	\includegraphics[width=0.12\textwidth,trim=150 10 150 30,clip]{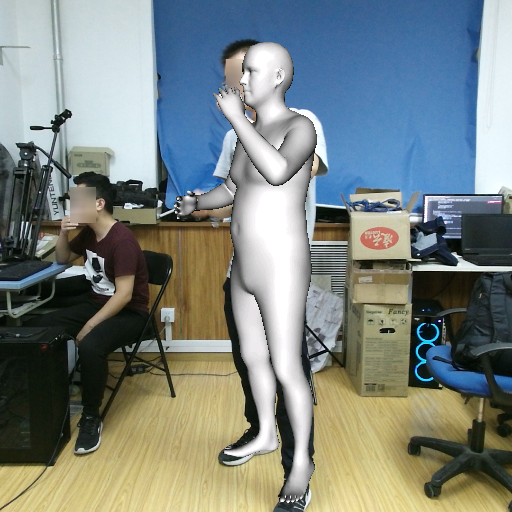}
	\hspace{-0.16cm}
	\includegraphics[width=0.12\textwidth,trim=150 10 150 30,clip]{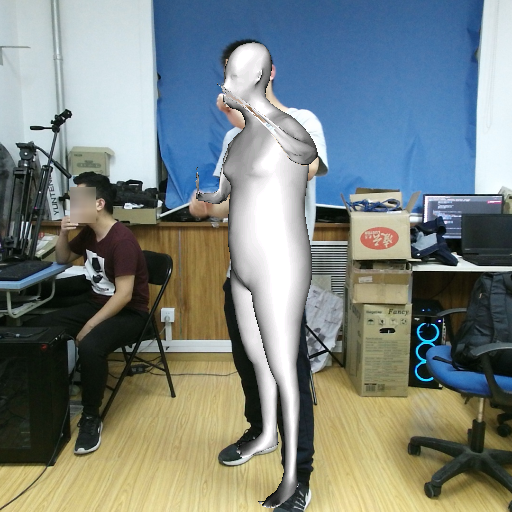}
	\hspace{-0.16cm}
	\includegraphics[width=0.12\textwidth,trim=150 10 150 30,clip]{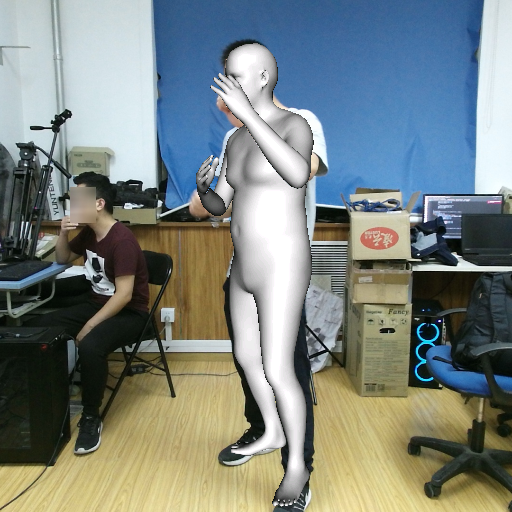}
	\hspace{-0.16cm}

	\begin{subfigure}{0.12\textwidth}
		\includegraphics[width=\textwidth,trim=150 10 150 30,clip]{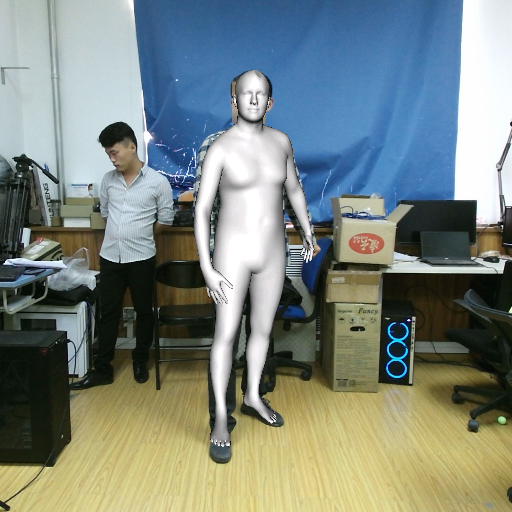}
		\caption{Our GCN}
	\end{subfigure}
	\hspace{-0.16cm}
	\begin{subfigure}{0.12\textwidth}
		\includegraphics[width=\textwidth,trim=150 10 150 30,clip]{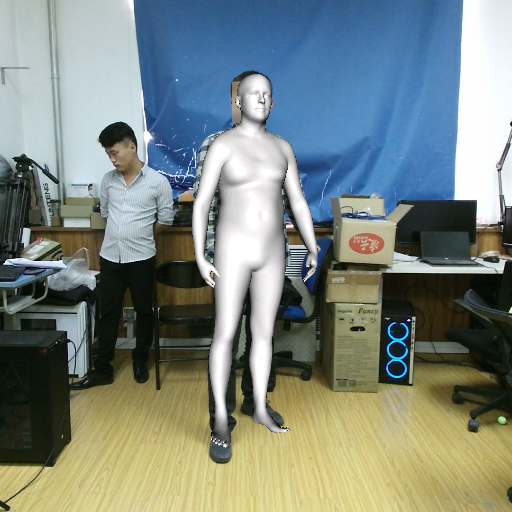}
		\caption{Linear Model}
	\end{subfigure}
	\hspace{-0.16cm}
	\begin{subfigure}{0.12\textwidth}
		\includegraphics[width=\textwidth,trim=150 10 150 30,clip]{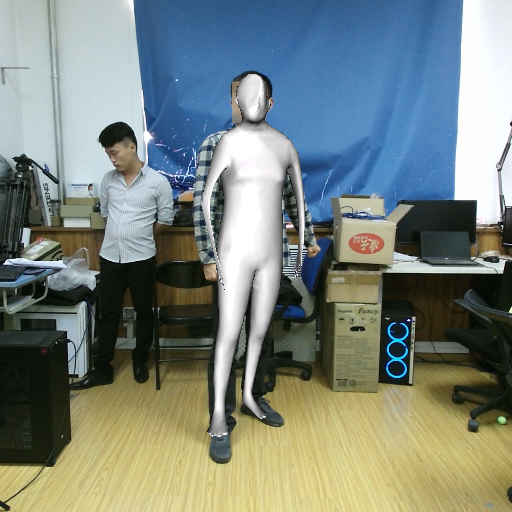}
		\caption{\scriptsize{GraphCMR GCN}}
	\end{subfigure}
	\hspace{-0.16cm}
	\begin{subfigure}{0.12\textwidth}
		\includegraphics[width=\textwidth,trim=150 10 150 30,clip]{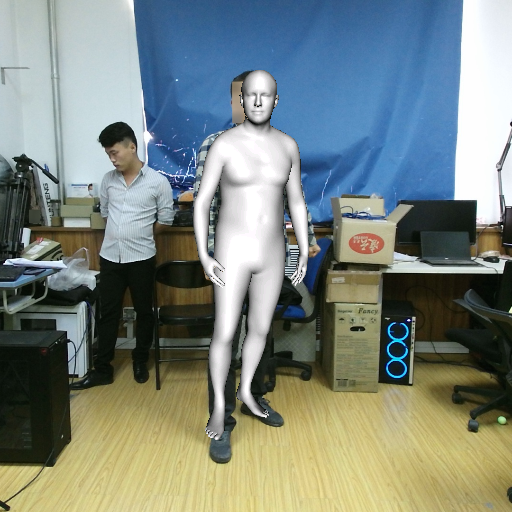}
		\caption{\scriptsize{GraphCMR Lin}}
	\end{subfigure}
	\hspace{-0.16cm}
	\caption{Comparisons on the model-based reconstruction. We show the results of our GCN decoder (a), linear model to predict the SMPL parameters (b), GCN results of GraphCMR (c) and final results of GraphCMR (d).}
	\label{fig:smpl}
\end{figure}

For model-based reconstruction results, we compare our GCN results with linear regressor~\cite{DBLP:conf/cvpr/HMR18} and GraphCMR~\cite{DBLP:conf/cvpr/graphcmr19}. We employ the same input and backbone network for all the methods. The mean per joint position error (MPJPE) and reconstruction error are used as the performance metrics. Table~\ref{tab:gcn} shows the experimental results. It can be clearly seen that our proposed GCN decoder obtains the lower estimation error comparing to other methods, which demonstrates the effectiveness of our GCN for body mesh representation. Fig.~\ref{fig:smpl} shows the visual results. As we perform graph convolution in spectral domain and have the normal and edge regularization, the mesh generated by our GCN is almost the same as the SMPL mesh. GraphCMR~\cite{DBLP:conf/cvpr/graphcmr19} generates over smoothing mesh, and needs another linear model to predict the SMPL parameters from the vertices predicted by GCN.

\begin{table}[htbp]
		\centering
		\begin{tabular}{c|c|c}
			\toprule
			Methods & MPJPE & Reconst. Error\\ 
			\midrule
			Linear Model & 50.058 & 42.367 \\
			GraphCMR GCN & 43.227 & 39.247 \\
			GraphCMR Linear & 40.276 & 36.023 \\
			GCN Decoder & \textbf{37.598} & \textbf{32.140} \\
			\bottomrule
		\end{tabular}
		\caption{Performance evaluation on model-based reconstruction.}
		\label{tab:gcn}
	\end{table}

\subsection{Evaluation on Our Collected Dataset}

\begin{figure}
    \centering
    
    \begin{subfigure}{0.115\textwidth}
        \includegraphics[width=\textwidth,trim=150 0 150 20,clip]{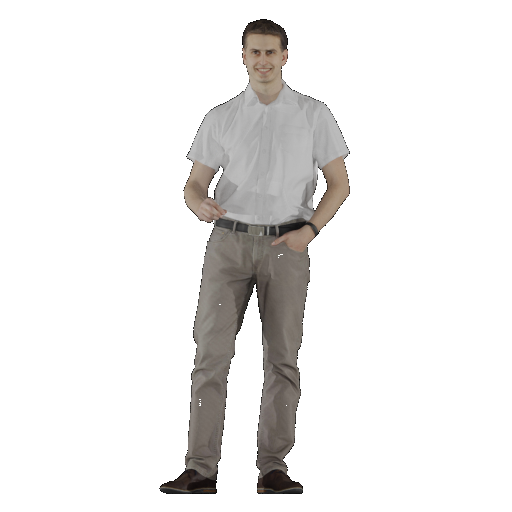}
        \caption{Input image}
    \end{subfigure}
    \hspace{-0.16cm}
    \begin{subfigure}{0.115\textwidth}
        \includegraphics[width=\textwidth,trim=150 0 150 20,clip]{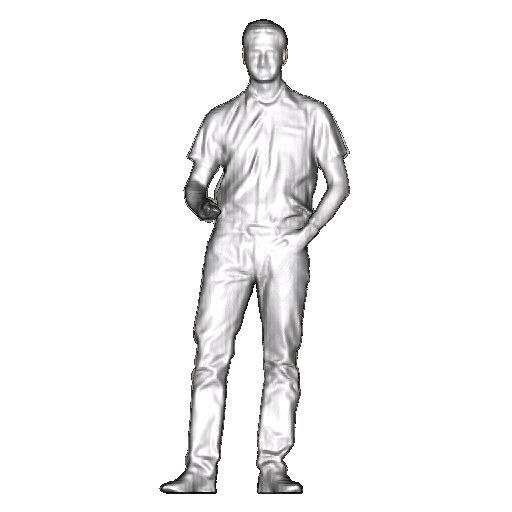}
        \caption{\scriptsize{implicit func}}
    \end{subfigure}
    \hspace{-0.16cm}
    \begin{subfigure}{0.115\textwidth}
        \includegraphics[width=\textwidth,trim=150 0 150 20,clip]{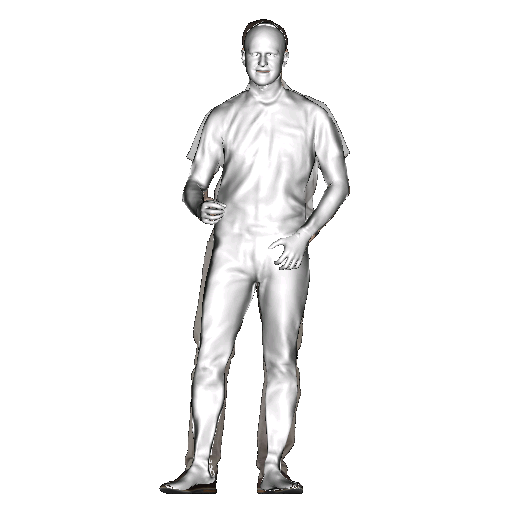}
        \caption{\scriptsize{with subdivide}}
    \end{subfigure}
    \hspace{-0.16cm}
    \begin{subfigure}{0.115\textwidth}
        \includegraphics[width=\textwidth,trim=150 0 150 20,clip]{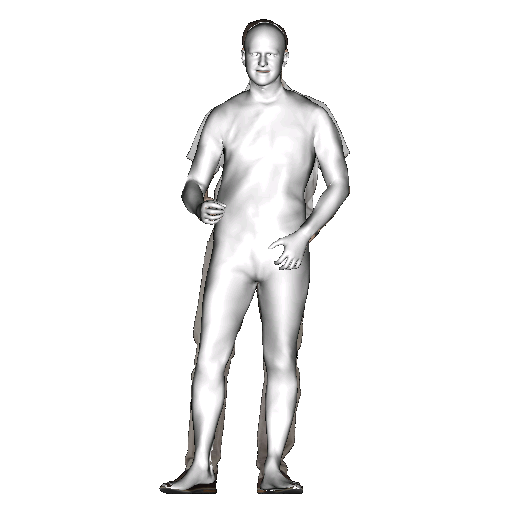}
        \caption{\scriptsize{w/o subdivide}}
    \end{subfigure}
    \hspace{-0.1cm}
    \caption{Comparison on the implicit registration results whether subdivide the SMPLX topology.}
    \label{fig:sub_results}
\end{figure}

Since either SMPL or SMPLX model has too few vertices to capture all clothing information during implicit registration stage, we subdivide its topology to generate more vertices. More specifically, for each face, we add the midpoint of each edge to subdivide every triangle into four facets. Fig~\ref{fig:sub_results} shows the effectiveness of subdivision.

\begin{table}
    \centering
    \begin{tabular}{c|c|c|c}
    \toprule
        Methods & Normal & P2S & Chamfer \\ 
    \midrule
        SMPLicit~\cite{DBLP:conf/cvpr/smplicit21} & 0.029 & 2.245 & 2.055 \\
        PIFuHD\,(our implementation) & 0.007 & 1.023 & 1.053 \\
        Ours & \textbf{0.005} & \textbf{0.969} & \textbf{0.965} \\
    \midrule
        *Inference results & 0.026 & 1.101 & 1.097 \\
        *Refined by Chamfer loss & 0.013 & 0.977 & 0.975 \\
        *Refined by implicit loss & 0.018 & 1.010 & 1.016 \\
    \bottomrule
    \end{tabular}
    \caption{Reconstruction Performance evaluation on our collected dataset. Same settings and notations as in Table~\ref{tab:evaluation on deephuman}.}
    \label{tab:evaluation on reconstruction}
\end{table}

We compare our proposed topology-preserved human reconstruction approach against the conventional Chamfer distance-based registration method used in IPNet~\cite{DBLP:conf/eccv/ipnet20} and PIFuHD~\cite{DBLP:conf/cvpr/pifuhd20}. Note that PIFuHD does not make their training code publicly available. Since the coordinate space of PIFuHD in human reconstruction is inconsistent with our collected scans, we re-implement PIFuHD for evaluation. Due to the limited number of high resolution scans having collected, the latent feature for GCN decoder cannot generalize well. Therefore, we further refine our GCN output to get correct human pose. 

\begin{table}[htbp]
    \small
    \setlength{\tabcolsep}{0.7mm}{
    \centering
    \centering
    \begin{tabular}{c|c|c|c}
    \toprule
        Methods & Marching Cube($512^3$) & Optimization & Total \\ 
    \midrule
        Chamfer Registration & 58.2s & 140.3s & 198.5s \\
        Ours & - & 27.0s & 27.0s \\
    \bottomrule
    \end{tabular}
    \caption{Comparison on computational time.}
    \label{tab:registration time}
    }
\end{table}

\begin{figure*}[ht]
    \centering

    \includegraphics[width=0.115\textwidth,trim=150 10 150 30,clip]{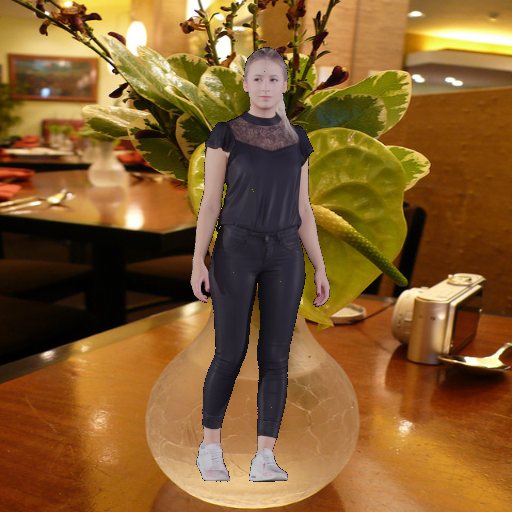}
    \hspace{-0.16cm}
    \includegraphics[width=0.115\textwidth,trim=150 10 150 30,clip]{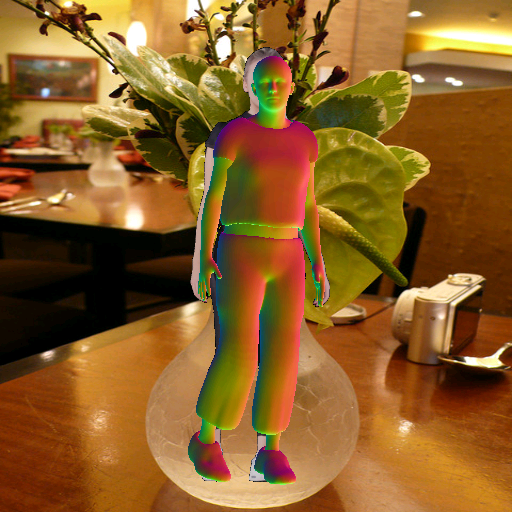}
    \hspace{-0.16cm}
    \includegraphics[width=0.115\textwidth,trim=150 10 150 30,clip]{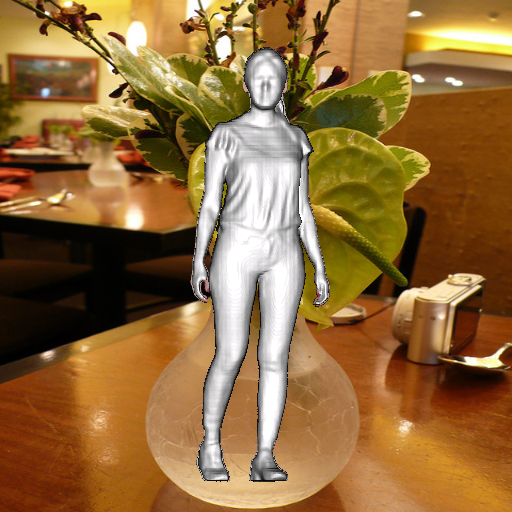}
    \hspace{-0.16cm}
    \includegraphics[width=0.115\textwidth,trim=150 10 150 30,clip]{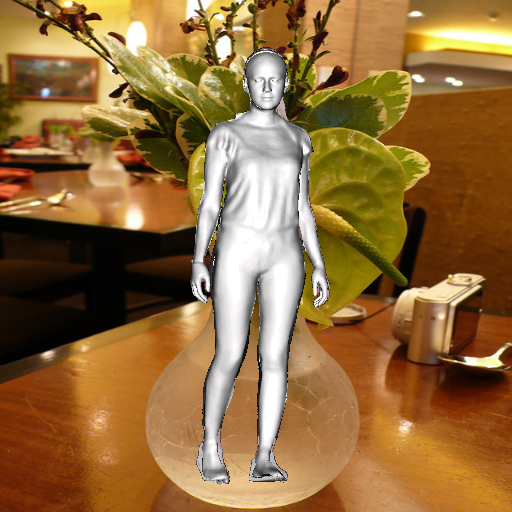}
    \hspace{-0.16cm}
    \includegraphics[width=0.115\textwidth,trim=150 10 150 30,clip]{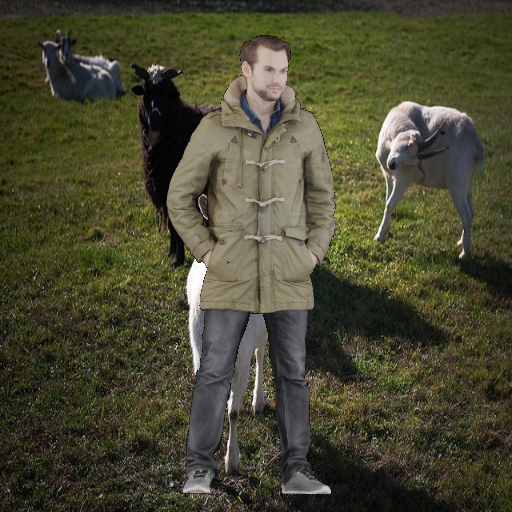}
    \hspace{-0.16cm}
    \includegraphics[width=0.115\textwidth,trim=150 10 150 30,clip]{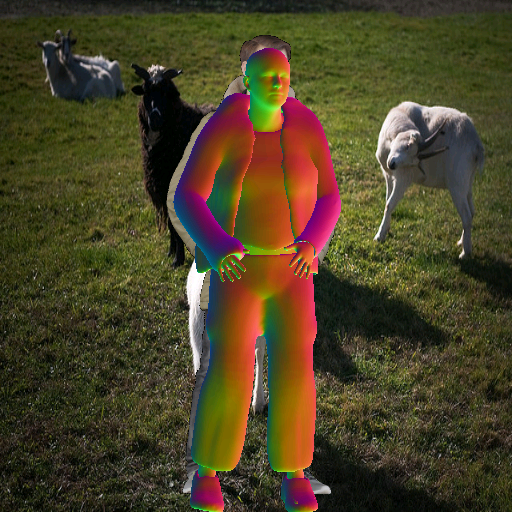}
    \hspace{-0.16cm}
    \includegraphics[width=0.115\textwidth,trim=150 10 150 30,clip]{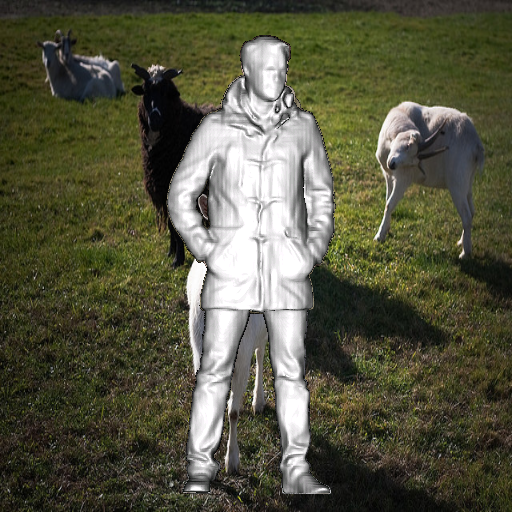}
    \hspace{-0.16cm}
    \includegraphics[width=0.115\textwidth,trim=150 10 150 30,clip]{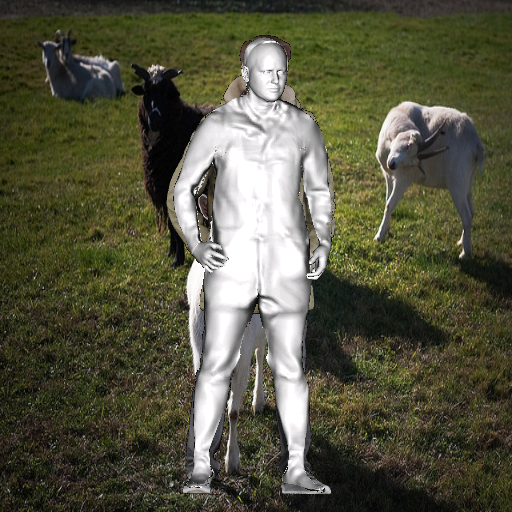}
    \hspace{-0.16cm}
    
    \begin{subfigure}{0.115\textwidth}
        \includegraphics[width=\textwidth,trim=150 10 150 30,clip]{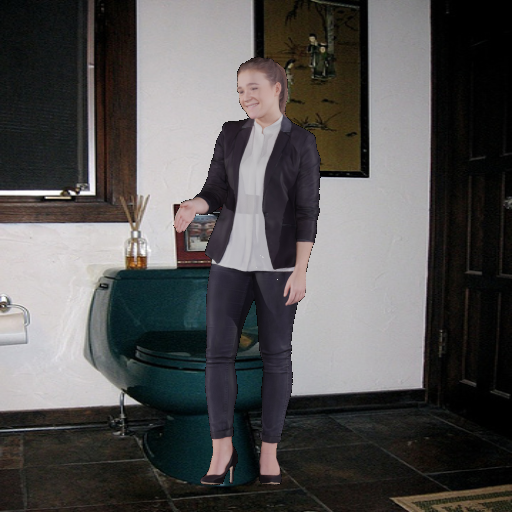}
        \caption{Input image}
    \end{subfigure}
    \hspace{-0.16cm}
    \begin{subfigure}{0.115\textwidth}
        \includegraphics[width=\textwidth,trim=150 10 150 30,clip]{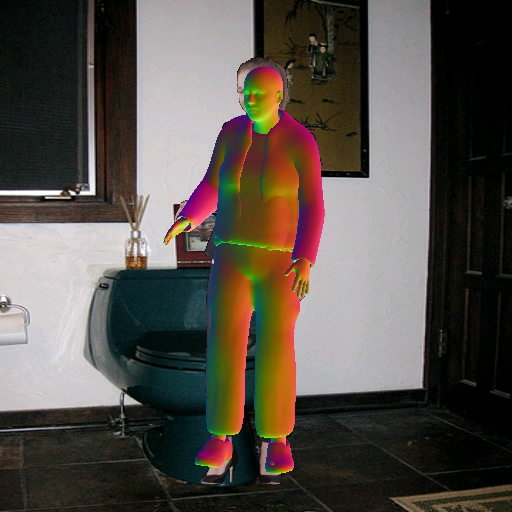}
        \caption{SMPLicit~\cite{DBLP:conf/cvpr/smplicit21}}
    \end{subfigure}
    \hspace{-0.16cm}
    \begin{subfigure}{0.115\textwidth}
        \includegraphics[width=\textwidth,trim=150 10 150 30,clip]{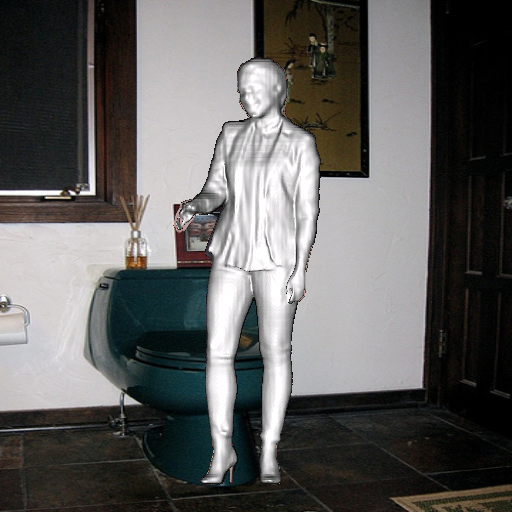}
        \caption{\scriptsize{implicit func}}
    \end{subfigure}
    \hspace{-0.16cm}
    \begin{subfigure}{0.115\textwidth}
        \includegraphics[width=\textwidth,trim=150 10 150 30,clip]{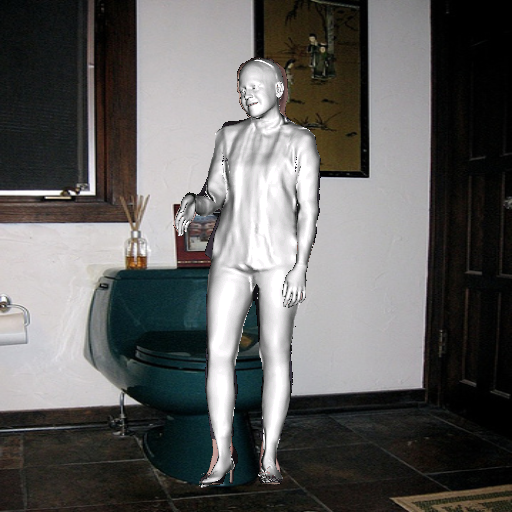}
        \caption{implicit loss}
    \end{subfigure}
    \hspace{-0.16cm}
    \begin{subfigure}{0.115\textwidth}
        \includegraphics[width=\textwidth,trim=150 10 150 30,clip]{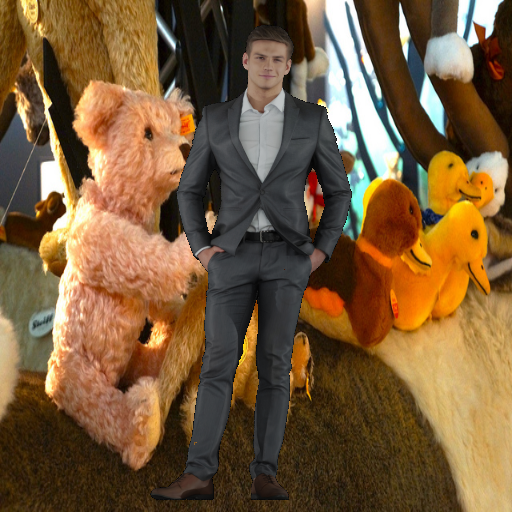}
        \caption{Input image}
    \end{subfigure}
    \hspace{-0.16cm}
    \begin{subfigure}{0.115\textwidth}
        \includegraphics[width=\textwidth,trim=150 10 150 30,clip]{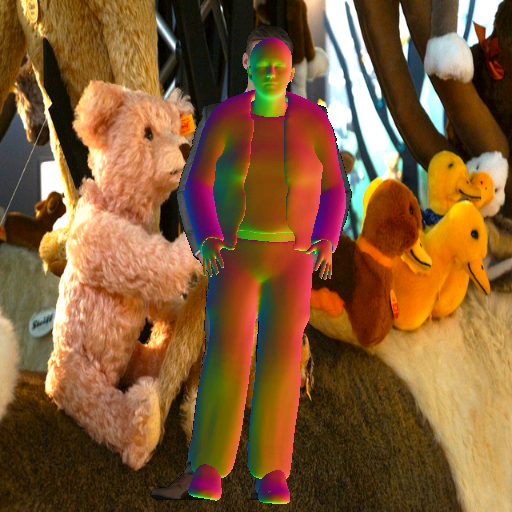}
        \caption{SMPLicit~\cite{DBLP:conf/cvpr/smplicit21}}
    \end{subfigure}
    \hspace{-0.16cm}
    \begin{subfigure}{0.115\textwidth}
        \includegraphics[width=\textwidth,trim=150 10 150 30,clip]{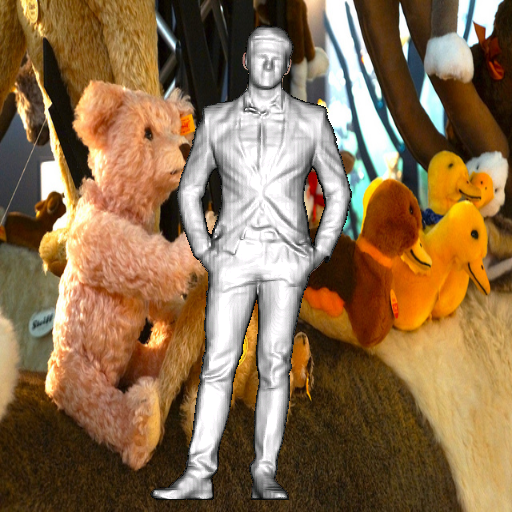}
        \caption{\scriptsize{implicit func}}
    \end{subfigure}
    \hspace{-0.16cm}
    \begin{subfigure}{0.115\textwidth}
        \includegraphics[width=\textwidth,trim=150 10 150 30,clip]{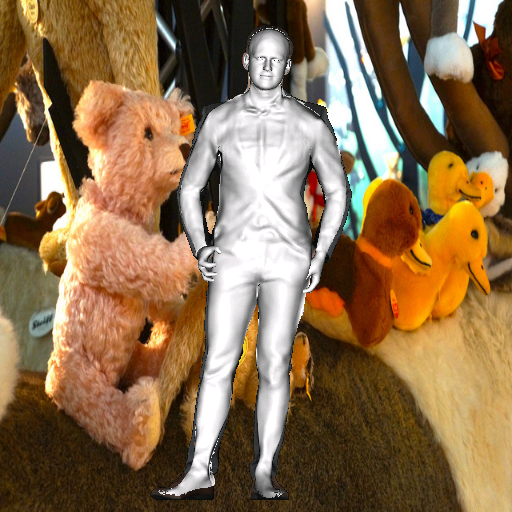}
        \caption{implicit loss}
    \end{subfigure}
    \hspace{-0.16cm}
    \caption{Results on our collected dataset. We show the results of SMPLicit~\cite{DBLP:conf/cvpr/smplicit21} (b,f), our proposed implicit function results (c,g) and implicit registration results (d,h), respectively.}
    \vspace{-0.2in}
    \label{fig:registration_results}
\end{figure*}

\begin{figure*}
    \centering
    \begin{subfigure}{0.11\textwidth}
        \includegraphics[width=\textwidth,trim=150 30 150 30,clip]{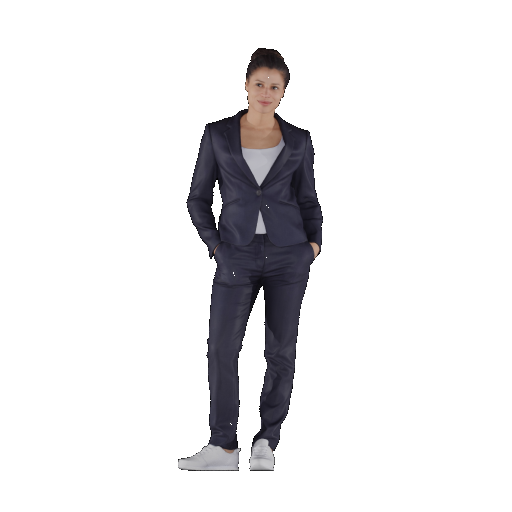}
        \caption{Input image}
    \end{subfigure}
    \hspace{-0.16cm}
    \begin{subfigure}{0.11\textwidth}
        \includegraphics[width=\textwidth,trim=150 30 150 30,clip]{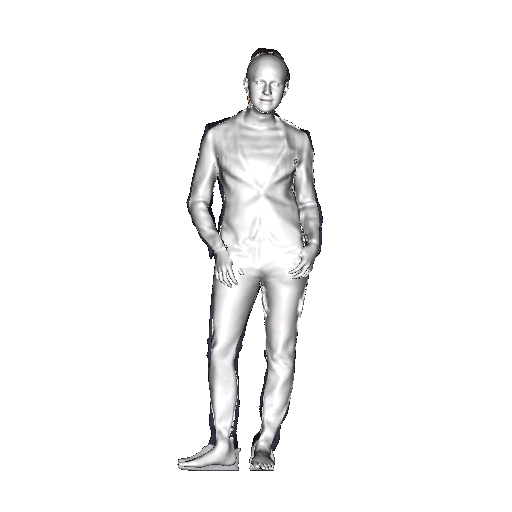}
        \caption{Recovered}
    \end{subfigure}
    \hspace{-0.16cm}
    \begin{subfigure}{0.11\textwidth}
        \includegraphics[width=\textwidth,trim=150 30 150 30,clip]{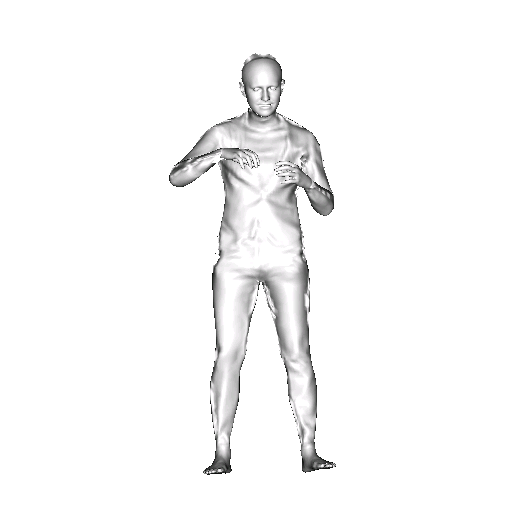}
        \caption{Reposed}
    \end{subfigure}
    \hspace{-0.16cm}
    \begin{subfigure}{0.11\textwidth}
        \includegraphics[width=\textwidth,trim=150 30 150 30,clip]{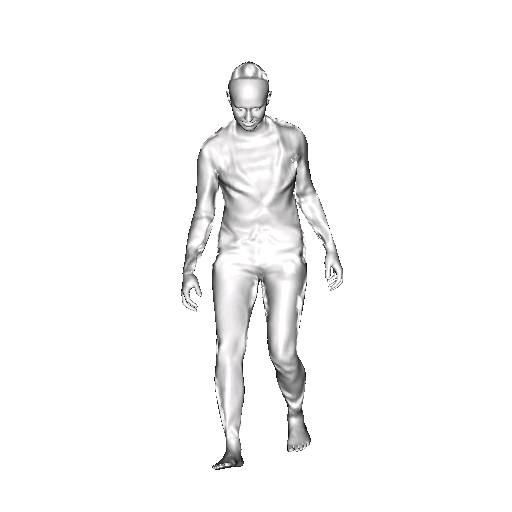}
        \caption{Reposed}
    \end{subfigure}
    \hspace{-0.16cm}
    \begin{subfigure}{0.11\textwidth}
        \includegraphics[width=\textwidth,trim=150 30 150 30,clip]{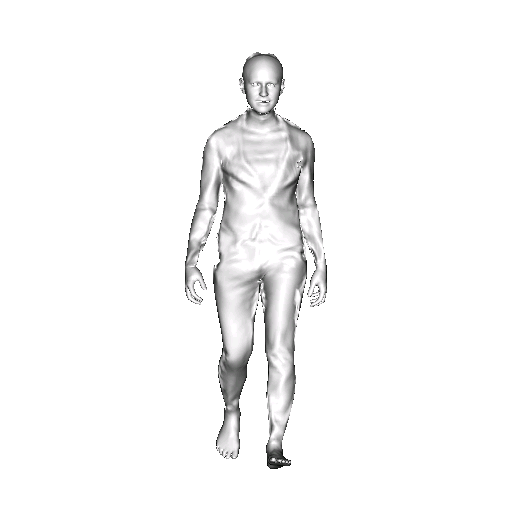}
        \caption{Reposed}
    \end{subfigure}
    \hspace{-0.16cm}
    \begin{subfigure}{0.11\textwidth}
        \includegraphics[width=\textwidth,trim=150 30 150 30,clip]{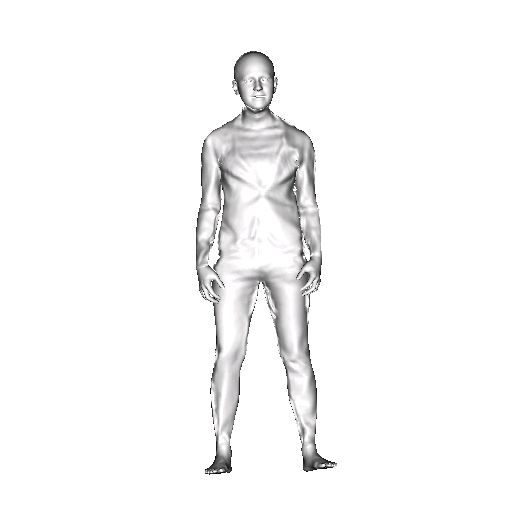}
        \caption{Reposed}
    \end{subfigure}
    \hspace{-0.16cm}
    \begin{subfigure}{0.11\textwidth}
        \includegraphics[width=\textwidth,trim=150 30 150 30,clip]{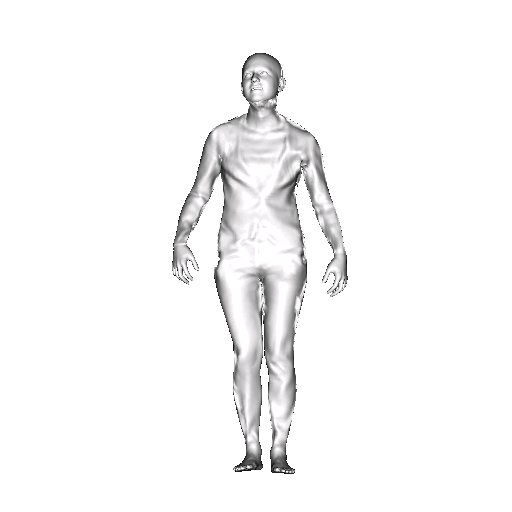}
        \caption{Reposed}
    \end{subfigure}
    \hspace{-0.16cm}
    \begin{subfigure}{0.11\textwidth}
        \includegraphics[width=\textwidth,trim=150 30 150 30,clip]{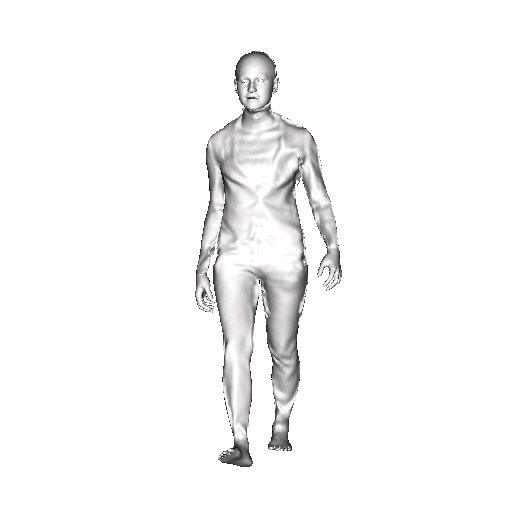}
        \caption{Reposed}
    \end{subfigure}
    \hspace{-0.1cm}
    \caption{Visual results of reposing the recovered mesh.}
    \vspace{-0.2in}
    \label{fig:reposed}
\end{figure*}

We use the same performance
metrics described above to evaluate our approach. As shown in Table~\ref{tab:evaluation on reconstruction}, our implicit function network performs comparable with PIFuHD. The mesh predicted by GCN decoder with the same topology as SMPLX model, which can be effectively refined by our proposed implicit registration scheme. It can be seen that the proposed approach performs comparable with conventional Chamfer distance-based method. Fig.~\ref{fig:registration_results} shows the reconstruction results. Due to the limited training data, SMPLicit~\cite{DBLP:conf/cvpr/smplicit21} cannot represent all kinds of clothing and lack of details. Since there are no ground-truth SMPLX parameters for face and hand in AMASS dataset, we employ the traditional optimization-based approach~\cite{DBLP:conf/cvpr/SMPLX19} to capture the hand and face for better visualization. By taking advantage of the fixed topology, we can easily animate the recovered mesh with the arbitrary poses, as shown in Fig.~\ref{fig:reposed}.

As depicted in Table~\ref{tab:registration time}, the speed of our implicit registration process is about seven times faster than the conventional Chamfer distance-based method. This is because the proposed implicit loss is efficient to compute while the Chamfer loss is computational intensive requiring to find the nearest neighbors. Moreover, our method does not require to extract the mesh by marching cube~\cite{DBLP:conf/siggraph/Marchingcubes87}, which saves the extra computational time.

\section{Conclusions}\label{sec:conc}
We proposed a novel topology-preserved human reconstruction approach to bridge the gap between model-based and model-free human reconstruction. The presented end-to-end neural network simultaneously predicts the pixel-aligned implicit surface and the explicit mesh surface built by graph convolutional neural network. Furthermore, we suggest the efficient implicit registration method to refine the neural network output in implicit space. We have conducted the evaluation on DeepHuman and our collected high resolution human dataset. The encouraging experimental results showed that our proposed approach is able to effectively recover the accurate mesh model while preserving its topology. 

{\small
\bibliographystyle{ieee_fullname}
\bibliography{egbib}
}

\end{document}